\documentclass[journal]{IEEEtran}

\ifCLASSINFOpdf

\else

\fi
\usepackage{color}
\usepackage{algorithm}  
\usepackage{algpseudocode}  
\usepackage{amsfonts}
\usepackage{algorithm} 
\usepackage{amsmath}
\usepackage{graphicx}
\usepackage{caption}
\usepackage[caption=false,font=normalsize,labelfont=sf,textfont=sf]{subfig}
\usepackage[numbers,sort&compress]{natbib}
\usepackage{multirow}

\begin{document}

\title{ Adversarial Masking Contrastive Learning for vein recognition}

\author{Huafeng Qin,
       Yiquan Wu,
    ~Mounim A. El-Yacoubi, Jun Wang, Guangxiang Yang
\thanks{Y. Wu, H. Qin, and G. Yang are with the Chongqing Engineering Laboratory of intelligent perception and block-chain, School of Computer Science and Information Engineering, Chongqing Technology and Business University, Chongqing 400067, China (e-mail: wuyiquan\_cq@163.com).}

\thanks{M. A. El-Yacoubi is with SAMOVAR, Telecom SudParis, CNRS, University Paris-Saclay, 91011 Paris, France (e-mail: mounim.el yacoubi@telecomsudparis.eu).}

\thanks{ Jun Wang is with the School of Information and Control Engineering, China University of Mining and Technology, Xuzhou, P.R. China (e-mail:jrobot@126.com).}

\thanks{Manuscript received June XX, 2023; revised XXXX XX, 201X. This work was supported in part by the National Natural Science Foundation of China (Grant Nos. 61976030, 62072061, and U20A20176), the Scientific Innovation 2030 Major
Project for New Generation of AI (Grant No. 2020AAA0107300), the Fellowship of China Post-Doctoral Science Foundation (Grant No. 59676651E), and the
Science Fund for Creative Research Groups of Chongqing Universities (Grant No. CXQT21034, Grant Nos. KJQN201900848 and KJQN201500814). }}

\markboth{Journal of \LaTeX\ Class Files,~Vol.~14, No.~8, August~2015}%
{Shell \MakeLowercase{\textit{et al.}}: Bare Demo of IEEEtran.cls for IEEE Journals}

\maketitle

\begin{abstract}
Vein recognition has received increasing attention due to its high security and privacy. Recently, deep neural networks such as Convolutional neural networks (CNN) and Transformers have been introduced for vein recognition and achieved state-of-the-art performance. Despite the recent advances, however, existing solutions for finger-vein feature extraction are still not optimal due to scarce training image samples. To overcome this problem, in this paper, we propose an adversarial masking contrastive learning (AMCL) approach, that generates challenging samples to train a more robust contrastive learning model for the downstream palm-vein recognition task, by alternatively optimizing the encoder in the contrastive learning model and a set of latent variables. First, a huge number of masks are generated to train a robust generative adversarial network (GAN). The trained generator transforms a latent variable from the latent variable space into a mask space. Then, we combine the trained generator with a contrastive learning model to obtain our AMCL, where the generator produces challenging masking images to increase the contrastive loss and the contrastive learning model is trained based on the harder images to learn a more robust feature representation. After training, the trained encoder in the contrastive learning model is combined with a classification layer to build a classifier, which is further fine-tuned on labeled training data for vein recognition. The experimental results on three databases demonstrate that our approach outperforms existing contrastive learning approaches in terms of improving identification accuracy of vein classifiers and achieves state-of-the-art recognition results. 

\end{abstract}


\IEEEpeerreviewmaketitle

\section{Introduction}

\IEEEPARstart{W}{ith} the rapid development of information technology and connected devices, privacy and security have gained unprecedented importance in our lives. Traditional verification methods, such as passwords, PINs, smart cards, plastic cards, tokens, and keys, cannot meet individual’s demands. For example, passwords and PINs are easy to forget and can be stolen or guessed; cards, tokens, and keys can be misplaced, forgotten, or duplicated; and magnetic cards can be destroyed.  To solve these problems, numerous higher security and privacy identification methods have been investigated. In recent decades, biometrics has received increasing attention and gradually been applied in different scenarios. Biometric traits used for recognition can be divided into two main categories:  physiological characteristics (e.g. facial \cite{r1}, fingerprint \cite{r2}, iris \cite{r3}, and vein \cite{r4}\cite{Qin2015_ICONIP} recognition), and  behavioral characteristics (e.g. gait \cite{r5} and voice \cite{r6} recognition). Physiological characteristics employed to verify personal identify include face and fingerprint recognition that have been widely applied in areas such as Information security, access control, law enforcement, smart cards and surveillance systems. Such biometrics technologies, however, are challenged by issues such as privacy leakage and spoofing attacks \cite{r7}. Face patterns, for instance, can be captured by cameras without users' awareness, and fake faces can be employed to deceive face recognition systems, even those using 3D face recognition \cite{erdogmus2013spoofing}. Similarly, fingerprints are easily left behind on touched objects \cite{zhang2012fingerprint} and are easy to replicate by a mold of the fingerprint with different materials, including gelatin, latex, and silicone. To solve these issues, researchers have investigated the use of the distribution information of vein vessels concealed beneath the body's skin for verification. Compared to surface patterns such as face and fingerprint, vein recognition has the following advantages: 1) High stability: The vein patterns are located beneath the skin and are not easily influenced by external conditions; 2) High security: It is difficult for others to obtain a person's vein pattern without users' consent, ensuring thereby high security; and 3) Liveness attribute: The blood and tissues in our body have different absorption rates under near-infrared light. When such a light passes through our body, the camera can record the projection distribution information of vein vessels. Thanks to these advantages, vein recognition has received special focus over the last years.

Vein recognition is a challenging task because the vein image quality is affected by many factors, e.g. environment illumination, light scatter,  environment temperature and user behavior. Therefore, there are some variations among the collected images from same hand/finger, which may increase the within-class distance, resulting in the degradation of recognition performance. To achieve robust recognition, some works \cite{r17}, \cite{r53} have designed various handcrafted descriptors such as Gabor filters and curvature to extract vein texture features, harnessed for matching to perform verification. However, the distortion, rotation and translation of vein vessels can cause mismatching between two images of the same finger. Compared to handcrafted approaches, machine learning based approaches can automatically learn discriminating features. Especially, recent deep learning methods have shown their ability to infer robust feature representation, which allowed them to achieve state-of-the-art performance in computer vision and object detection tasks. It is no surprising then that deep learning models such as Convolutional neural network (CNN) and Transformers have been become, recently, an increasing application trend in vein biometrics tasks. Deep learning based approaches, nonetheless, require large amounts of labeled training data to optimize the models' millions of parameters. Labeling such datasets, however, can be time-consuming, expensive, or even impractical for certain domains or niche applications, which limits the learning capacity of deep learning based methods. In addition, small dataset sizes generally lead to model overfitting. Likewise, in the vein recognition task, it is difficult to collect huge data from each user because of the following facts. 1) To protect their privacy, users usually reluctant to provide more samples; 2) Collecting large amounts of samples is time consuming and expensive. Also, it is more convenient that users are required to provide only few samples. In practice, vein datasets \cite{r5}, \cite{r8,r9}, \cite{r14,r15}, \cite{r40} typically include no more than 20 images for each object. To address the data scarcity issue, researchers have introduced GANs for data augmentation. The works \cite{r39}, \cite{r40}, \cite{r44}, for example, utilize GANs to enlarge palm vein datasets for classifier training. For the same purpose, GANs were proposed to generate palmprint images \cite{r45}, and  finger vein images \cite{r46}, \cite{r41}.  However, as GANs also need a large training dataset to generate effectively augmented data, directly employ GANs for data augmentation lead to a conflict. As a matter of fact, GANs trained with limited data tend to overfit the data distribution, which results in poor performance in terms of generating diverse data. 

Recently, unsupervised representation learning such as contrastive learning methods have received special attention as it greatly reduces the cost of collecting a large volume of labeled data to train deep networks. These schemes have been highly successful in natural language processing, e.g., as shown by GPT \cite{brown2020language} and BERT \cite{devlin2018bert}. Several recent studies \cite{chen2020improved, r49,chen2021shot} brought them into visual feature representation learning with promising results. During the contrastive learning, the objective is to maximize the similarity between positive sample pairs (pairs of similar samples) while minimizing the similarity between negative sample pairs (pairs of dissimilar samples). Contrastive learning allows for efficient utilization of data by creating positive and negative pairs, which enabling the model to capture rich and meaningful representations without the need for extensive manual labeling.

 Inspired by this success, we propose, in this paper, an adversarial masking contrast learning approach (AMCL) for vein recognition. First, we randomly generate large mount of masks, based on which a GAN is effectively trained to model the distribution of the mask space. Then, the trained generator from GAN is combined with the contrastive learning model to constitute our AMCL, where the adversarial targets are the input sets of latent variables for the generator and the  contrastive learning model. During the adversarial training process, the  contrastive learning model aims to learn the feature representation of input images by minimizing the contrastive loss, and the latent variable sets are selected to generate challenging samples to increase the loss of the contrastive learning model. After training, the resulting encoder in the contrastive learning model is capable of extracting more effective features. Finally, the classification layer with softmax is combined with the resulting encoder to form a classifier, which is further fine-tuned with labeled training  vein data to improve palm vein recognition accuracy.  

 The main contributions of our work are as follows:

1) We  propose a novel contrastive learning approach, named adversarial masking contrastive learning (AMCL), for vein recognition. The proposed AMCL generates a large amount of challenging samples to train the encoder in the contrastive learning model. The classification model, consisting of the trained encoder and softmax, is capable of learning more robust features to improve recognition accuracy.

2) To perform palm-vein data augmentation, we explore a GAN to learn the mask distribution for image augmentation. Then, an adversarial framework is proposed to jointly optimize the target encoder training and the latent variable search.

3) We conducted extensive experiments on three public palm vein datasets to verify the performance of the proposed method. The experimental results demonstrate that our method outperforms existing contrastive approaches and achieves a significant recognition performance improvement w.r.t all vein  classifiers.

The rest of this work is organized as follows: In section II, we further introduce related work. The proposed method is detailed in section III. The experiments carried out to verify the performance of the proposed method are detailed in section IV. The last section summarizes this work.
\section{Related work}
Vein recognition is still a challenging task because vein image quality can be affected by several factors during the capturing process. The various methods investigated to ensure robust recognition can be broadly divided into the following four categories.


\begin{figure*}[!t]
	\centering
    
	\includegraphics[scale=0.5]{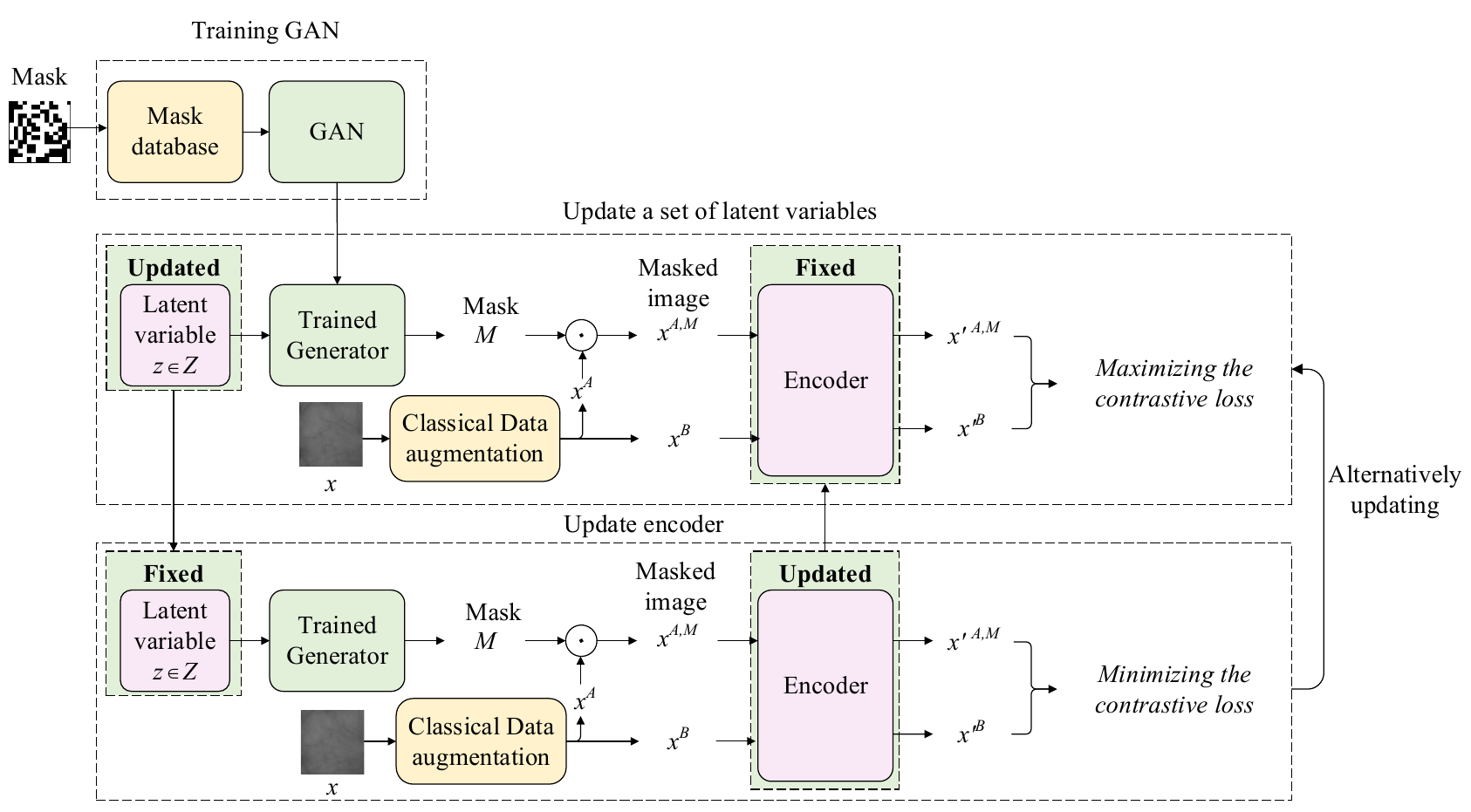}%
		\label{fig_first_case}
   
	\caption{Framework of AMCL}
	\label{fig_sim}
\end{figure*}

\subsubsection{Handcrafted vein recognition approaches}
In this category, the approaches design descriptors based on humans' prior knowledge.  Typical works include valley detection-based methods \cite{r8,r9,r10,r11,r12,r13,r14}, line detection-based methods \cite{r15,r16,r17}, and local descriptor-based methods \cite{r19,r20,r21,r22}. For example, researchers have found that vein patterns show valley-like shapes, and built mathematical models to detect such valleys for vein segmentation. In \cite{r9}, the vein patterns are extracted  by computing the local maximum curvature in the vein cross section. Subsequently, in \cite{r48}, the mean curvature method was used to detect valley-like structures. To  further improve recognition performance, Qin, et al. \cite{r14} proposed  a region growth model for vein feature extraction. In \cite{r13},  vein patterns are segmented by tracking lines at different positions. Line detection-based methods treat the vein patterns as linear features and extract the vein features by detecting such straight lines. Gabor filters are representative approaches and have been used for vein linear feature detection \cite{r17}, \cite{r53}. For local descriptor-based methods, local feature descriptors such as Local Binary Patterns (LBP)\cite{r20}, effective Local Binary Patterns (ELBP) \cite{r22}, and Discriminative Binary Coding (DBC) \cite{r54} have been proposed for extracting vein features.
\subsubsection{Shallow learning based vein recognition approaches}
 Shallow learning-based methods can be viewed as a neural network with one or two layers for feature representation learning and include Principal Component Analysis (PCA) \cite{r23}, Kernel Principal Component Analysis (KPCA) \cite{r24}, Kernel Entropy Component \cite{r25}, sparse representation  \cite{r29, r30}, etc. For example, in \cite{r26}, PCA and metric learning are combined to learn a KNN classifier for vein recognition. Work \cite{r27} employs linear discriminant analysis to extract low-dimensional discriminating features. Work \cite{r28} combines hybrid-based local phase quantization and SVM to improve vein recognition accuracy. In \cite{r29}, sparse representation is employed for learning vein features. Similarly, in \cite{r30}, a mutual similarity representation classification (MSRC) is introduced to enhance vein recognition performance.
\subsubsection{Deep learning based vein recognition approaches}
  Compared to shallow learning approaches, deep learning methods can learn high-level features and abstract representations from raw data by stacking multiple layers. Various deep learning based approaches \cite{r18}, \cite{r31,r32,r33, r34,r35,r36,r37,r38,[74],[112],[119]} have been proposed recently for vein recognition. For example, Das et al. \cite{[74]} made one of the first attempts to employ the CNN for finger-vein recognition. In \cite{r34}, a palm vein recognition approach is investigated by combining the features from Binarized Statistical Image Features (BSIF) descriptor and CNN model.  In \cite{r35},  an architecture called SingelNet with fewer convolutional layers was proposed to extract features for palm vein recognition. In \cite{r37}, a pre-trained ViT is fine-tuned on a palm vein dataset to extract more robust features for palm vein recognition. Qin et al. \cite{r38} proposed a label-enhanced multiscale vein Transformer for palm vein recognition by combining a Transformer and label enhancement model.
\subsubsection{Data augmentation-based vein recognition approaches}
Traditional deep learning methods cannot fully leverage their performance due to the limited amount of data in palm vein datasets. Data augmentation is a powerful solution introduced to address this issue. Typical methods such as Generative Adversarial Networks (GANs) have been proposed to generate images for vein data augmentation. For example, Qin et al. \cite{r39} proposed a multi-scale multi-direction GAN  to learn patch distributions within a single image for data synthesis. In 2022, Ou et al. \cite{ou2022gan} proposed a GAN-based framework that can generate arbitrary-pattern vein images based on the binary vein texture, with the capability to augment the training data with new vein classes. The work \cite{r41} combined GAN with Triplet loss to generate vein data to improve the learning capability of the classifier. Similarly, a fully convolutional Generative Adversarial Network (FCGAN) \cite{r42} was proposed to generate finger-vein images. The work \cite{r43} proposed to enlarge data through vertical flipping and conventional data augmentation approaches. The authors considered also a fusion loss by incorporating the classification loss and the metric learning loss to enhance the discrimination of deep features.

\begin{table*}
	\begin{center}
		\caption{Architecture of generator}
		\label{tab1}
		\begin{tabular}{ l l l l l l}
			\hline
			Operator & Output & Kernel & KSize & Stride & Padding\\
			\hline
			Input noise & 128×1×1 & - & - & - & - \\
                ConvTranspose-ReLU-BatchNorm & 2048×4×4 & 2048 & 4 & 1 & 0 \\
                ConvTranspose-ReLU-BatchNorm & 1024×8×8 & 1024 & 4 & 2 & 1 \\
                ConvTranspose-ReLU-BatchNorm & 512×16×16 & 512 & 4 & 2 & 1 \\
                ConvTranspose-ReLU-BatchNorm & 256×32×32 & 256 & 4 & 2 & 1 \\
                ConvTranspose-Tanh & 1×64×64 & 1 & 4 & 2 & 1 \\
			\hline 
		\end{tabular}
	\end{center}
\end{table*}

\begin{table*}
	\begin{center}
		\caption{Architecture of discriminator}
		\label{tab2}
		\begin{tabular}{ l l l l l l}
			\hline
			Operator & Output & Kernel & KSize & Stride & Padding\\
			\hline
			Input image & 1×64×64 & - & - & - & - \\
                Conv-ReLU-BatchNorm & 32×16×16 & 32 & 8 & 4 & 2 \\
                Conv-ReLU-BatchNorm & 64×8×8 & 64 & 4 & 2 & 1 \\
                Conv-ReLU-BatchNorm & 128×4×4 & 128 & 4 & 2 & 1 \\
                Conv-Sigmoid & 1×1×1 & 1 & 4 & 1 & 0 \\
			\hline 
		\end{tabular}
	\end{center}
\end{table*}

\section{METHOD}
Contrastive leaning is an unsupervised learning scheme capable of learning robust feature representations to improve the performance of downstream tasks. It has achieved promising performance in NLP \cite{bhattacharjee2022text} and computer vision \cite{r49}.  The robustness of contrastive learning relies on constructing a collection of positive examples and negative examples that are sufficiently hard to discriminate against positive queries when their representations are self-trained. The data augmentation policy for samples synthesising  plays a critical role in terms of learning the effective representation for contrastive leaning \cite{r49}. In most works \cite{r49, he2020momentum, chen2020improved}, classical data augmentation approaches, such as image resizing, random color jittering, random horizontal flip, random grayscale conversion, and blur augmentation, are employed for generating positive and negative examples. Recently, Masked image modeling (MIM) has been successfully applied in NLP \cite{devlin2018bert} and computer vision \cite{he2022masked,xie2022simmim}. MIM randomly masks a portion of input signals and tries to predict these masked signals, which enables the prediction model to learn better the underlying mask distribution in a self training way. Inspired by this success, we introduce a masking operator to enlarge data for contrastive learning and propose an adversarial masking contrastive learning for vein recognition. First, we  randomly produce a large mount of masks, based on which a GAN is trained to learn the mask distribution space. After training, taking a latent vector as input, the generator outputs a mask. The mask generation operations with discrete parameters (i.e. masked size and masking ratio) are transformed into a continuous latent space as well as the differentiable mechanism for selecting operations. Second, we combine the trained generator with the contrastive learning model to propose a joint generative and contrastive learning framework for vein recognition. During training, the contrastive learning aims to learn the representations by maximizing the agreement between differently augmented examples of the same data example via a contrastive loss. By contrast, a latent variable set as a whole is learned to increase the training loss of the contrastive learning model through generating adversarial samples. In other words, the GAN provides online data augmentation for contrastive learning, while the contrastive learning model learns invariant features for vein representation. Finally, the resulting encoder is obtained from the trained contrastive learning model and employed for downstream palm vein classification tasks.

\subsection{Generative adversarial network}
Due to limited training data, directly training GAN to provide augmentation data for contrastive learning may result in overfitting. Some works \cite{xie2022simmim,he2022masked} show that randomly masking an image is helpful for deep learning models to learn more robust feature representations. In our work, we propose a GAN to learn the distribution space based on a large number of masked samples. Our GAN consists of a generator and a discriminator, detailed next.
\subsubsection{Generator}
The generator consists of 5 convolutional layers. In each layer, the number of neurons gradually increases to meet the resolution of the generated images. The first 4 convolutional layers include convolution, BatchNorm, and LeakyReLU function, while the last layer contains convolution and Tanh function. As shown in Table I, detailing the generator architecture, the first layer takes the latent vector as input and outputs 2,048 feature maps with size of 4 $\times$ 4. For the last four layers, the fractionally-stridden convolutions can be treated as a pixel upsampling block. The resolution of the input feature map increases two times, while the embedded dimension reduces to its half in each layer, except for the last layer. In the last layer, the size of the output is the same as that of the target samples, and the dimension of the feature map is reduced to 64 $\times$ 64.

\subsubsection{Discriminator}
The discriminator consists of 4 convolutional layers, which gradually reduce the size of the feature map and increase the embedding dimension. The first three convolutional layers are composed of convolution, BatchNorm, and LeakyReLU, while the last convolutional layer contains convolution and Sigmoid function. As shown in Table II, illustrating the discriminator architecture, the input image resolution is 1$\times$64 $\times$ 64. After forwarding three convolutional layers, the resolution is reduced gradually to 4 $\times$ 4. Finally, the last layer outputs the probability value of the input image being real or fake.

\subsubsection{Adversarial loss}
The idea of classical GANs is to establish a minimax game between two players: The discriminator is responsible for distinguishing real training data from synthetic images, while the generator aims to fool the discriminator. During the training process, the generator $G$ takes the latent variable $z$ as input to generate images $x'=G(z)$, while the discriminator takes real images $x$ and generated images $x'$ as inputs and learn to distinguish the real images from the synthesized ones. The objective of the GAN is defined by Eq.(\ref{eq1}).
\begin{equation}
\begin{aligned}
    L_GAN (D,G)=&E_{x~P_{data}(x)} logD(x)+\\
                &E_{z~P(z) }log[1-D(G(z))]
                \label{eq1}
\end{aligned}
\end{equation}
where the $P_{data}(x)$ is data distribution and $P(z)$ is the model distribution. 
\subsubsection{Sample Masking}
Generally, the GAN is trained on a large dataset to learn the real data distribution and then generates samples from such a distribution space. Because of limited  data, however, we cannot train a robust GAN for sample generation. To solve this problem, we generate a huge number of masks based on crafted approaches, which will detail in Section IV-B. Then, a robust GAN is trained to learn the mask distribution. Finally, a large mount of masks are generated based on the trained generator $G_*$ by Eq.(\ref{eq2}).

\begin{equation}
\begin{aligned}
       M=G^*(z)
        \label{eq2}
\end{aligned}
\end{equation}

By combining the mask with the vein image, we can generate plenty of occluded images for contrastive learning training, according to the following Eq.(\ref{eq3}).. 

\begin{equation}
\begin{aligned}
       x^{M}=x \odot  M
        \label{eq3}
\end{aligned}
\end{equation}
where $\odot$ denotes a Hadamard product and $x^{M}$ is the occluded image of $x$. 

\subsection{Contrastive Loss}
\subsubsection{Encoder}
In general, the original images have high dimensionality, so it is difficult to effectively compute the contrastive loss based on such images for contrastive learning model training. As deep learning models have shown robust feature representations capacity,  the high level deep features are extracted from original images by deep learning models for contrastive learning. In recent years, various deep learning based classifiers have been employed for vein recognition and achieved state-of-the-art recognition performance. The encoder can be obtained by removing the classification layer from existing classifiers such as ResNet, FV-CNN, PV-CNN, FVRAS-Net, Lightweight CNN, and ViT.  Based on the resulting encoder, the original image $x$ is embedded into a low dimensional vector by Eq.(\ref{eq4}).:
\begin{equation}
    x'=E(x)
     \label{eq4}
\end{equation}
where $E$ denotes the encoder.
\subsubsection{Contrastive loss}
Contrastive learning aims to learn the representations of data by maximizing the similarity between positive pairs (examples that should be similar) and minimizing the similarity between negative pairs (examples that should be dissimilar). Therefore, the quantity and quality of positive and negative samples are crucial for the performance improvement. Assume that there are $ N $ examples from minibatch. In general, the original samples are subject to transformation by the classic augmentation approaches such as random cropping, random color distortions, and random Gaussian blur to generate the positive samples. Usually, two samples are produced from each original image. As a result, there are 2$N$ generated images, which constitutes $N$ positive pairs. The negative examples is explicitly produced. Instead, for an augmented image, we treat the other 2$(N-1)$ augmented examples within a minibatch as negative examples. Specifically, give a batch with \textit{N} images $x={x_i }$, $i=1,...,N$ and two classic data augmentation sets $A$ and $B$ (rotation, scaling, flipping, cropping, color jittering) are applied to each images to obtain two sets of different views $x^A$ and $x^B$, which are then fed into an encoder $E$ to obtain corresponding feature vectors $x'^A$ and $x'^B$ . The contrastive learning \cite{xie2022simmim} is defined as Eq.(\ref{eq5}).:
\begin{equation}
    L_{SimCLR}(x,E)=-log\dfrac{exp(S(x'^A_i,x'^B_i))}{\sum_{i\neq j} exp(S(x'^A_i,x'^B_j))}
     \label{eq5}
\end{equation}
where $S$ is the cosine similarity. In fact, Eq.(\ref{eq5}) tries to minimize the distance between the representations of the two augmented views (i.e., positive
samples) from the same image, and maximize the distance between the representations from different images. This enables the model to effectively capture the similarities and differences between samples, thereby further improving recognition performance.

In our work, we combine the SimCLR \cite{xie2022simmim}  objective with GAN and then follow the exact same pipeline. For example, we replace the $x'^A_i$ by $x'^{A,M}_i$, where  $x'^{A,M}_i=E(x^{A} \odot M)$, the  Hadamard product $\odot$ is performed between $x^A$ and the generated mask $M$, resulting in masked images $x^{A,m}$. Following equation Eq.(\ref{eq5}), we can derive our objective function as Eq.(\ref{eq6}).:
\begin{equation}
\begin{aligned}
         L_{SimCLR}(x,z, E,G^*) =-\dfrac{exp(S(x'^{A,M}_i,x'^{B}_i))}{\sum_{i\neq j} exp(S(x'^{A,M}_i,x'^B_j))} \\
         =-log\dfrac{exp(S(E(x^{A}_i \odot G^*(z)), E(x^B )))}{\sum_{i\neq j} exp(S(E(x^{A}_i \odot G^*(z)),E(x^{B}_j))}
          \label{eq6}
\end{aligned}
\end{equation}

\subsection{Adversarial network}
Based on different input latent variable $z$, the generator in Eq.(\ref{eq2}) can synthesize different masks which are combined with original images to obtain occluded images by the Hadamard product. In \cite{r49, chen2020improved}, a finite number of samples, e.g one or two positive samples for each original image, are randomly generated for contrastive learning model training in an offline way. However, the image pair generation in these works is not directly related to the target task, i.e. the reduction of the contrastive loss. As a consequence, the randomly selected synthesized sample pairs may not be representative for the contrastive learning task, which degrades the encoder generalization in the contrastive learning model. To tackle this issue, we introduce an adversarial framework, consisting of a contrastive learning model and a GAN, to alternatively optimize the encoder in the contrastive learning model and a set of latent variables, thereby alleviating the limitation of mere random generation of input samples. Specifically, a latent variable set as a whole is learned to increase the contrastive loss through generating challenging samples, while the encoder in the contrastive learning model learns more robust features from the resulting harder examples to improve the generalization.  
\subsubsection{Adversarial loss function}
 The weights of encoder $E$ are defined as $\vartheta$. The input of the generator is a set of latent variables $\mathbb{Z}$. The adversarial contrastive learning loss for the training is denoted by Eq.(\ref{eq7}).:
\begin{equation}
    \vartheta^*, \mathbb{Z}^*=arg\underset{\vartheta}{min}\underset{ \mathbb{Z}}{max}((\underset{z\in \mathbb{Z}, x \in \mathbb{X}} {\mathbb E}[L_{SimCLR}(x,z, E_{\vartheta},G^*)])
     \label{eq7}
\end{equation}
where the encoder weights ${\vartheta}$  are  optimized to obtain the representation of each query in current mini batch. Similarly, the latent variables in $\mathbb{Z}$ can be treated as free variables of composite function $E_{\vartheta}(x, x \odot G^*(\cdot)$ and are optimized for a given ${\vartheta}$. During the training process, the positive pairs and negative pairs are fed to the encoder model for training by minimizing the contrastive loss. To optimize the parameter ${\vartheta}$, $ G^*(\cdot))$ generates hard images with given latent variables to increase the contrastive learning loss, which may lead, however, to the collapse of the inherent meanings of images. To address this drawback, we introduce the $cosine$ similarity as a regularization term to control the quality of synthesised images. The objective loss function is defined, accordingly, by Eq.(\ref{eq8}).

\begin{equation}
\begin{aligned}
    \vartheta^*, \mathbb{Z}^*=arg\underset{\vartheta}{min}\underset{ \mathbb{Z}}{max}(\underset{z\in \mathbb{Z}, x \in \mathbb{X}} {\mathbb E}[L_{SimCLR}(x,z, E_{\vartheta},G^*)\\ + \lambda cosine(x^A, x^A \odot G^*(z))])
     \label{eq8}
    \end{aligned}
\end{equation}

  \begin{algorithm}[t]
  \caption{ Joint Training of the contrastive learning model and a latent variable set search.}
  \label{alg:Framwork}
  \begin{algorithmic}[1]
    \Require
       The initial set of latent variables, $\mathbb {Z}=[z_1,z_2,...,z_k]$; Batch of examples $\mathbb{X}={x}^N_{i=1}$; The encoder $E_\vartheta$; The trained generator $G^*$; 
    \Ensure
      $\vartheta^*$, $\mathbb{Z}^*$;
    \State For 1 $\leq$ e $\leq$ $epoch$;
       \quad \label{code:fram:extract}
    \State  \quad Taking the latent variable set $\mathbb{Z}$ as the input of the trained generator $G_{*}$ to generate masks, which are combined with original samples in $\mathbb{X}$ to obtain the augmented (occluded) images by Eq.(3). Then we combine augmentated images with the original samples  $\mathbb{X}$ to obtain sample set $\Omega$; We randomly select $N$ sample pairs from $\Omega$ to construct mini-batches.
    \State \quad For 1 $\leq$ $t_1$  $\leq$ $T_1$;
    \label{code:fram:add}
    \label{code:fram:trainbase}
    \State \quad \quad Update $\vartheta(t+1)$ according to Eq. (13);
    \State \quad end
     \label{code:fram:add}
    \State \quad For 1 $\leq$ $t_2$  $\leq$ $T_2$;
        \label{code:fram:add}
   \State  \quad \quad {Update $z(t+1)$ according to Eq.(9)}; 
    \label{code:fram:classify}
    \State \quad \quad {Update set $\mathbb {Z}(t+1)$}.
     \State \quad end
     \State end
  \end{algorithmic} \label{algorithm1}
\end{algorithm}

\subsection{Optimization}
Similar to existing adversarial algorithms, it is difficult to find the optimal solution $(\vartheta^*, \mathbb{Z}^*)$ of Eq.(\ref{eq8}). Usually, the optimization problem of adversarial learning can be transformed into solving two separate optimization problems. Specifically, the gradient descent method is often used to update $\varphi$ and the gradient ascent method is used to solve $\vartheta$. By alternately updating $\vartheta$ and $\vartheta$, we can solve the optimization problem in Eq.(\ref{eq8}).  For example, the optimization process of $\varphi$ can be transformed into Eq.(\ref{eq9}):
\begin{equation}
\begin{aligned}
    \vartheta^*=arg\underset{\vartheta}{min}(\underset{z\in \mathbb{Z}, x \in \mathbb{X}} {\mathbb E}[L_{SimCLR}(x,z, E_{\vartheta},G^*)\\ + \lambda cosine(x^A, x^A \odot G^*(z))])
     \label{eq9}
    \end{aligned}
\end{equation}

 Vanilla SGD with a learning rate of $\alpha$ and a batch $N$ is employed to solve the optimal problem in Eq.(\ref{eq9}). The updating procedure for each batch is defined as 

\begin{equation}
\begin{aligned}
    \vartheta(t+1) =  \vartheta(t) -\alpha \frac{1}{N} \sum_{n=1}^{N} \nabla_\vartheta [L_{SimCLR}(x,z, E_{\vartheta},G^*)\\ + \lambda cosine(x^A, x^A \odot G^*(z))]
     \label{eq10}
    \end{aligned}
 \end{equation}
As $cosine(x, x \odot G^*(z))$ is dependent with variable $\vartheta$, so the Eq.(\ref{eq10}) is converted to Eq.(\ref{eq11}).

\begin{equation}
\begin{aligned}
    \vartheta(t+1) =  \vartheta(t) -\alpha \frac{1}{N} \sum_{n=1}^{N} \nabla_\vartheta [L_{SimCLR}(x^A,z, E_{\vartheta},G^*)]
     \label{eq11}
    \end{aligned}
 \end{equation}

To accelerate convergence, an average gradient over $N$ instances is usually computed to reduce gradient variance during the training procedure. However, the randomly generated training sample pairs are not related to the contrastive learning model training, which may degrade generalization. To solve this problem, a set of occluded images are generated by a GAN to increase the training loss of the target network, which leads to a minmax problem to self-train the network. Such a scheme may guide the GAN to generate more challenging samples, so as to improve the representation capacity of the target task. The objective function is defined by the following maximization problem.

\begin{equation}
\begin{aligned}
    \mathbb{Z}^*=arg\underset{ \mathbb{Z}}{max}(\underset{z\in \mathbb{Z}, x \in \mathbb{X}} {\mathbb E}[L_{SimCLR}(x,z, E_{\vartheta},G^*)\\ + \beta cosine(x^A, x^A \odot G^*(z))])
     \label{eq12}
    \end{aligned}
\end{equation}

 Gradient ascent can be employed to solve the optimization problem with learning rate $\beta$ and the latent set $ \mathbb{Z}$ is updated by Eq.(\ref{eq13}).

\begin{equation}
\begin{aligned}
    z(t+1) =  z(t) +\beta  \frac{1}{N} \sum_{n=1}^{N} \nabla_z [L_{SimCLR}(x,z, E_{\vartheta},G^*)\\ + \lambda cosine(x^A, x^A \odot G^*(z))]
     \label{eq13}
    \end{aligned}
 \end{equation}

 The optimization of  Eq.(\ref{eq13}) tends to push the positive samples farther from the queries in the current minibatch, thus resulting in challenging  positives closely tracking the change of the updated network. The adversarial learning process is shown in Algorithm 1.

\begin{figure}[!t]
	\centering
    \vspace{-0.7cm}
	\includegraphics[scale=0.5]{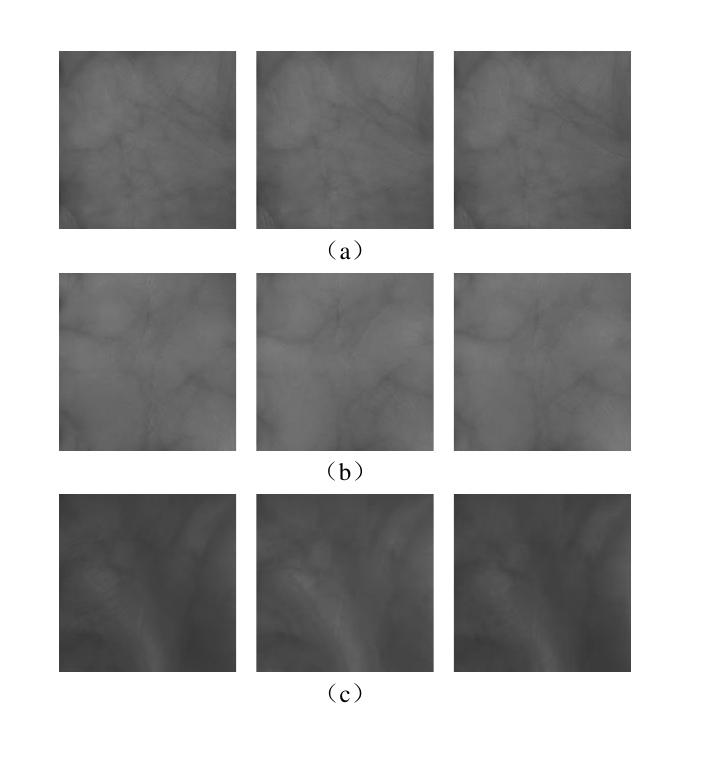}%
		\label{fig_second_case}
    \vspace{-0.6cm}
	\caption{ROI images from (a) dataset A, (b) dataset B and (c) dataset C.}
	\label{fig_2}
\end{figure}

\begin{figure*}[!t]
	\centering
	\includegraphics[scale=0.8]{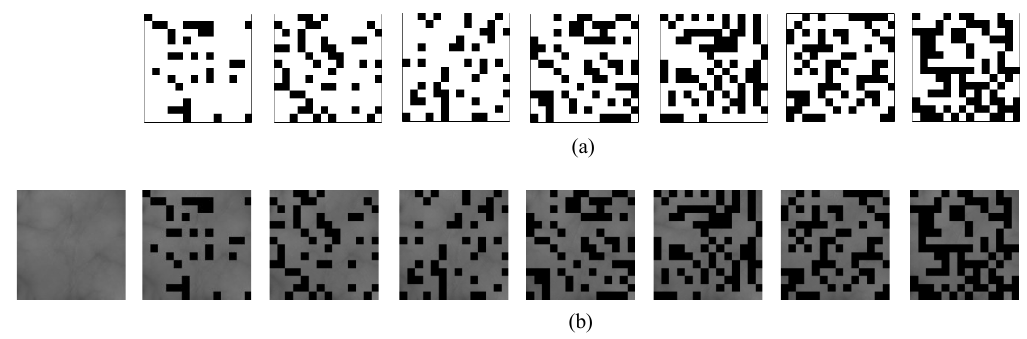}%
		\label{fig_second_case}
	\caption{Masked results. (a) Masks generated by different masking strategies using different mask ratios with 16-sized patches, and (b) Masked images with different masks.}
	\label{fig_3}
\end{figure*}
\section{EXPERIMENTS}
To estimate the performance of our approach,  we have carried out rigorous experiments on three publicly available palm vein datasets. To evaluate its effectiveness in terms of improving recognition accuracy, we compare the proposed approach with classic deep leaning models, namely ResNet\cite{resnet} 
as well as state-of-art vein classifiers, namely FV-CNN \cite{[74]}, PV-CNN \cite{r39}, FVRAS-Net \cite{[112]}, and LightWeight CNN \cite{[119]}. For sound comparison, we assess as well baseline contrastive learning methods, namely SimCLR\cite{xie2022simmim}, ADIOS \cite{r50} and VICReg\cite{r51}. All experiments are conducted on a computer with NVIDIA Tesla A100 GPUs and PyTorch.

\subsection{Datasets}
\subsubsection{Dataset A}: The CASIA palm vein dataset consists of 1200 (100 volunteers × 2 hands × 3 images × 2 sessions) palm vein images from 100 volunteers, with each providing two hands: left and right. The images were collected in two separate sessions, with 3 images captured from each palm at each session. Each image is encoded according to a 256 gray-scale distribution with a resolution of 200 $\times$ 200.

\subsubsection{Dataset B}: The VERA palm vein dataset contains 2200 (110 subjects × 2 hands × 5 images × 2 sessions) palm vein images from 110 volunteers, at two independent sessions. Each subject provides 20 images (5 images $\times$ 2 hands $\times$ 2 sessions) from the left and right hands. For each session, 5 images are captured from each palm. The resolution of each image is 200 $\times$ 200. 

\subsubsection{Dataset C}: The Tongji University palm vein dataset includes 12000 (300 subjects $\times$ 2 hands $\times$ 10 images $\times$ 2 sessions) palm-vein images from 300 volunteers, collected at two sessions with an interval of approximately 60 days. 10 images are captured from each palm in each session. The resolution of each image is 128 $\times$ 128. 
 
As the original images contain background areas that are not helpful for classification, we extract the Regions of Interest (ROI) from such images by approach \cite{qin2019iterative}. All ROI images are normalized to 64 $\times$ 64. Fig.\ref{fig_2} shows the ROI images from the three datasets.

\subsection{Experimental setting}
To validate the performance of the proposed method, we divide each database into two separate sets. For database A, we treat each palm as a class, obtaining thereby a total of 200 classes (100 volunteers $\times$ 2 palms). We employ the images from the first session for training and the images from the second session for testing. There are 6,000 images (200 hands $\times$ 3 images) in the training set and testing set, respectively. Similarly, the training set and testing sets comprise 2,200 images (220 hands $\times$ 5 images) for dataset B, and  6,000 images (300 volunteers $\times$ 2 palms $\times$ 10 images) for database C. 
 For image augmentation, we randomly generate 100,000 masks (as shown in Fig.\ref{fig_3}(a)) with $S \times S $ patch size in a wide range of masking ratios (20\%-80\%).  Here, the $S$ is set to 16 based on the work in \cite{xie2022simmim}. Then, a robust GAN is trained to learn the distribution of the mask space.  After training, the GAN is capable of generating any mask by taking a latent variable as its input. Finally, we combine the trained generator with the contrastive learning model for adversarial learning. During the adversarial training stage, a set of latent variables is fed to the generator to generate masks that are further combined with the original images to obtain masked images (as shown in Fig.\ref{fig_3}(b)). In the experiments, two synthesised images are generated for each original image based on classical data augmentation approaches and one of the augmented images is then subject to masking to obtain the corresponding masked image. The remaining augmented one and the resulting masked image from the same image constitute a positive pair, and two augmented images (masked images, synthesised images or mixed images) from different original images are treated as a negative pair. The resulting sample pairs are employed to challenge contrastive learning model. The contrastive learning model can learn a robust feature representation based on the generated sample pairs by minimizing the contrastive loss. After the adversarial training, the trained encoder in the contrastive learning model is used for downstream tasks such as vein recognition by adding a classification layer. 

\begin{table*}
	\begin{center}
		\caption{Recognition results of various methods on dataset A}
		\label{tab3}
		\begin{tabular}{ c c c c c c c c c c c }
			\hline
            \multirow{2}*{Classifier}{} & \multicolumn{2}{c}{Scratch} & \multicolumn{2}{c}{SimCLR} & \multicolumn{2}{c}{VICReg} & \multicolumn{2}{c}{ADIOS} & \multicolumn{2}{c}{Ours}   \\
            ~ & ACC & EER & ACC & EER & ACC & EER & ACC & EER & ACC & EER \\
			\hline
			ResNet          & 86.83 & 1.95 & 91.67 & 1.15 & 92.00 & 1.48 & 92.67 & 1.68 & 96.50 & 1.08\\
            FV-CNN          & 92.83 & 1.33 & 93.17 & 2.06 & 92.83 & 1.13 & 93.67 & 0.94 & 96.83 & 0.51\\
            PV-CNN          & 90.17 & 2.85 & 90.83 & 1.99 & 91.50 & 2.50 & 91.33 & 2.38 & 92.67 & 1.86\\
            FVRAS-Net       & 84.17 & 2.51 & 89.17 & 1.64 & 87.50 & 1.72 & 90.00 & 1.54 & 93.67 & 1.08\\
            Lightweight CNN & 86.67 & 1.30 & 90.83 & 1.33 & 87.00 & 1.34 & 91.33 & 1.36 & 95.50 & 1.18\\
			\hline 
		\end{tabular}
	\end{center}
\end{table*}

\begin{table*}
	\begin{center}
	\caption{Recognition results of various methods on dataset B}
		\label{tab4}
		\begin{tabular}{ c c c c c c c c c c c }
			\hline
            \multirow{2}*{Classifier}{} & \multicolumn{2}{c}{Scratch} & \multicolumn{2}{c}{SimCLR} & \multicolumn{2}{c}{VICReg} & \multicolumn{2}{c}{ADIOS} & \multicolumn{2}{c}{Ours}   \\
            ~ & ACC & EER & ACC & EER & ACC & EER & ACC & EER & ACC & EER \\
			\hline
			ResNet          & 88.36 & 1.75 & 93.45 & 0.85 & 95.65 & 1.37 & 96.36 & 1.43 & 97.91 & 0.81\\
            FV-CNN          & 93.09 & 1.27 & 97.36 & 0.91 & 95.55 & 1.25 & 96.82 & 0.81 & 98.27 & 0.71\\
            PV-CNN          & 91.36 & 2.56 & 95.09 & 1.29 & 94.73 & 1.20 & 94.45 & 1.11 & 96.64 & 0.97\\
            FVRAS-Net       & 89.36 & 1.90 & 91.09 & 1.37 & 92.27 & 1.20 & 92.18 & 1.28 & 97.00 & 0.82\\
            Lightweight CNN & 91.91 & 1.71 & 94.73 & 0.98 & 90.00 & 1.90 & 94.91 & 1.11 & 95.73 & 0.98\\
			\hline 
		\end{tabular}
	\end{center}
\end{table*}

\begin{table*}
	\begin{center}
		\caption{Recognition results of various methods on dataset C}
		\label{tab5}
		\begin{tabular}{ c c c c c c c c c c c }
			\hline
            \multirow{2}*{Classifier}{} & \multicolumn{2}{c}{Scratch} & \multicolumn{2}{c}{SimCLR} & \multicolumn{2}{c}{VICReg} & \multicolumn{2}{c}{ADIOS} & \multicolumn{2}{c}{Ours}   \\
            ~ & ACC & EER & ACC & EER & ACC & EER & ACC & EER & ACC & EER \\
			\hline
			ResNet          & 88.70 & 1.45 & 92.45 & 0.90 & 92.33 & 1.07 & 93.28 & 0.92 & 97.65 & 0.83\\
            FV-CNN          & 89.23 & 2.78 & 92.43 & 1.50 & 92.58 & 1.53 & 92.55 & 1.20 & 95.37 & 1.05\\
            PV-CNN          & 89.85 & 1.82 & 92.97 & 1.32 & 91.50 & 1.30 & 90.98 & 1.47 & 95.07 & 0.92\\
            FVRAS-Net       & 86.57 & 1.77 & 89.22 & 1.33 & 87.43 & 1.95 & 87.25 & 1.55 & 96.27 & 0.73\\
            Lightweight CNN & 92.80 & 1.00 & 93.27 & 0.91 & 88.70 & 1.37 & 93.43 & 0.97 & 95.45 & 0.79\\
			\hline 
		\end{tabular}
	\end{center}
\end{table*}

 \begin{figure*}[t]
	\centering{
   \begin{minipage}[b]
		{.3\linewidth}\centering
			\subfloat[]{\includegraphics[scale=0.35]{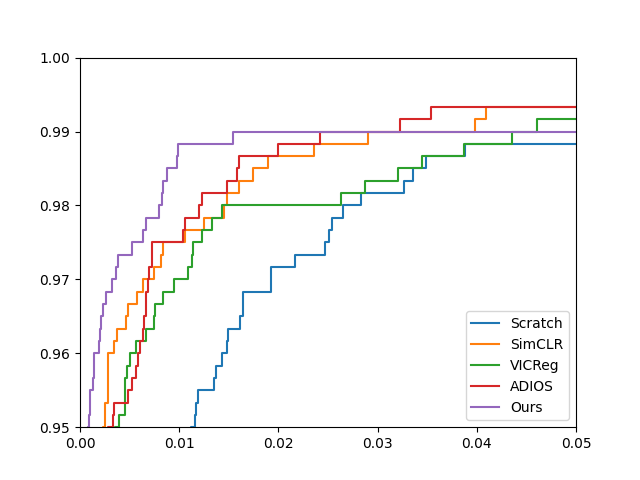}}\\
		\end{minipage}
	}
		\begin{minipage}[b]{0.3\linewidth}\centering
			\subfloat[]{\includegraphics[scale=0.35]{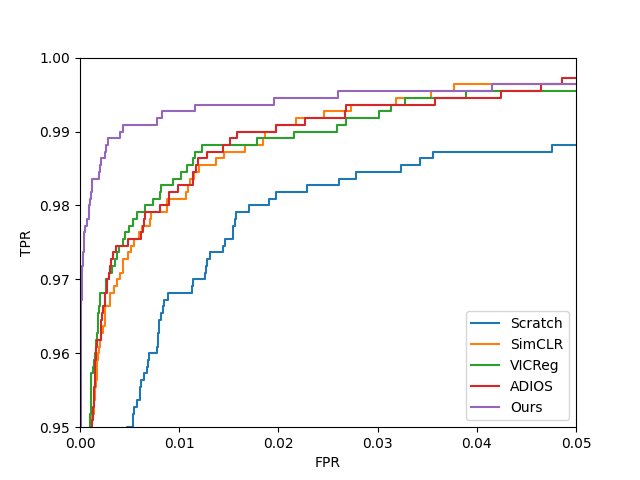}}\\
		\end{minipage}
	 	\begin{minipage}[b]{.3\linewidth}\centering
			\subfloat[]{\includegraphics[scale=0.35]{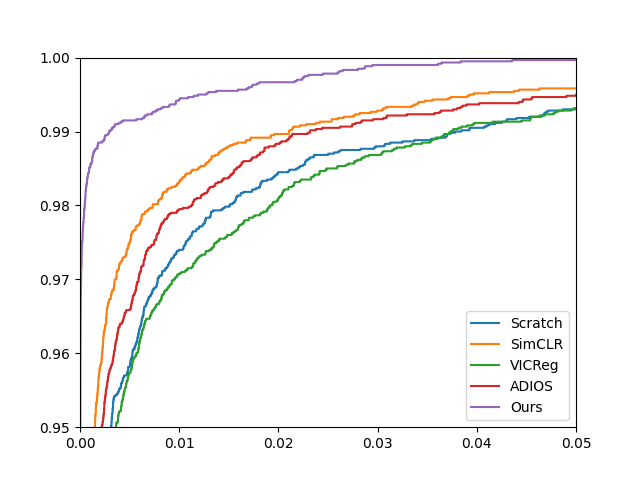}}\\
		\end{minipage}
	\caption{ROC curves of various approaches on (a) database A; (b) database B and (c) database C.}
	\label{fig4}
\end{figure*}
\subsection{Experimental results}
We have conducted extensive experiments on three public palm vein datasets to evaluate the performance of our approach. Concretely, we have implemented 6 classifiers, namely ResNet, FV-CNN, PV-CNN, FVRAS-Net,  and LightWeight CNN, for which we show the recognition results on the three databases. Additionally, we remove the classification layer of each classifier to obtain the corresponding encoder, which we train then based on our approach in adversarial way. After training, the classification layer with softmax is combined with the pre-trained encoder to constitute a classifier, which is further fine-tuned on labeled vein training images for recognition. In the same way, the six vein classifiers are pre-trained by the contrastive learning methods, i.e. SimCLR, VICReg and ADIOS. We report the results of all vein classifiers before or after pre-training for comparison. Also, we try to depict the ROC curves of all the classifiers trained by the four contrastive learning methods. However, as there are three ROC curves for each classifier on the three databases, 5 classifiers result in 15 curves. Therefore, we only report the ROC curve of the state of the art vein classifier, i.e. FVRAS-Net on the three databases for comparison.  Tables III, IV, and V show the recognition accuracy and Equal Error Rate (EER) of the these classifiers on dataset A, dataset B, and dataset C, respectively. Fig.\ref{fig4} displays the ROC curves of the different approaches.

The experimental results (Fig.\ref{fig4}, Table III, Table IV, and Table V) indicate that our method outperforms SimCLR, VICReg and ADIOS as it improves the recognition accuracy, and achieves the highest recognition accuracy and lowest EERs on the three datasets. For example, using our approach for pre-training, the classifiers achieve the lowest EER  (1.59\%, 1.14\% and 1.23\%) on database A, database B, and  database C.  Overall, the six classifiers achieve a significant improvement on identification accuracy after training based on existing contrastive learning approaches (i.e SimCLR, VICReg and ADIOS) as well as on our approach, which implies that the vein classifier benefits from the unsupervised contrastive learning framework. Such a good performance can be explained by the following facts: 1) the original deep learning based vein classifiers are directly trained on the labeled vein datasets. It is difficult, however, to collect large amounts of labeled vein data for model training. In general, the existing vein datasets include no more than 20 images from each class, which makes it virtually impossible to effectively train a deep learning model with millions of parameters. The learning capacity of deep learning based approaches is therefore not effectively exploited, which results in limited performance. 2) Contrastive learning models as self-trained deep networks are trained to distinguish the representation of positive samples from their negative counterparts. As a large number of samples are generated by data augmentation approaches such as cropping, color distortions, random blur, and masking, the contrastive learning models are capable of learning a robust feature representation to improve the performance of downstream tasks.  From the experimental results, we also observe that our approach outperforms the existing contrastive learning approaches, i.e. SimCLR, VICReg, and ADIOS, in terms of improving the recognition accuracy of vein classifiers. This may be attributed to the following facts: For existing approaches SimCLR and VICReg, the positive samples are randomly generated by classical data augmentation schemes such as cropping and rotation, and then input to a contrastive learning model during the whole training process. As the image pair generation is directly related to the contrastive learning model, the randomly synthesised image pairs may not be representative for target task training.  By contrast, our approach focuses on searching an optimal set of latent variables to generate masked images along with the training process, rather than synthesising new image pairs in an offline way. As the image pair generation process is guided by maximizing the loss of the contrastive learning, harder images may be generated to gradually train a more robust contrastive learning model during the adversarial training process. In other words, the automatic masking increases the challenging power of the randomly generated sample pairs by maximizing the contrast loss, so as to improve the robustness of the contrastive learning model.

\section{CONCLUSION}
In this paper, We have proposed a joint generative and contrastive learning framework for vein recognition, which combines  GAN  and contrastive learning to learn a robust vein classifier. First, a very large number of masks are generated to train a robust GAN for learning the mask distribution space. Second, the trained GAN is combined with the contrastive learning model to obtain our AMCL, which is trained in an adversarial way. Specifically, a set of latent variables is searched for generating challenging sample pairs to increase the loss of the contrastive learning model. The  contrastive learning model is able to learn a robust feature representation based on generated hard samples. Our experimental results on three public databases show that our method outperforms existing contrastive learning approaches in terms of improving the performance of vein classifiers and achieve state-of-the-art recognition accuracy.





\ifCLASSOPTIONcaptionsoff
  \newpage
\fi


\bibliographystyle{unsrt}
\bibliography{references}










\end{document}